\documentclass[10pt,twocolumn,letterpaper]{article}

\usepackage{cvpr}              
\usepackage[accsupp]{axessibility}  

\newcommand{\TODO}[1]{\textbf{\color{red}[TODO: #1]}}
\renewcommand{\TODO}[1]{}

\usepackage{microtype}

\renewcommand{\paragraph}[1]{\vspace{.5em}\noindent\textbf{#1.}}

\usepackage{soul}
\setuldepth{foobar}

\usepackage[T1]{fontenc}
\usepackage{graphicx}
\usepackage{amsmath,amssymb}
\usepackage{booktabs}
\usepackage{multirow}
\usepackage{xspace}
\usepackage{capt-of}
\usepackage{xcolor}

\newcommand{\bx}{\mathbf{x}}
\newcommand{\bz}{\mathbf{z}}
\newcommand{\bh}{\mathbf{h}}
\newcommand{\ba}{\mathbf{a}}
\newcommand{\bW}{\mathbf{W}}
\newcommand{\bb}{\mathbf{b}}
\newcommand{\R}{\mathbb{R}}
\newcommand{\cD}{\mathcal{D}}

\newcommand{\paper}{\textbf{GeoSAE}\xspace}

\definecolor{adred}{HTML}{c0392b}
\definecolor{sexblue}{HTML}{2980b9}
\definecolor{geneteal}{HTML}{17becf}

\definecolor{cvprblue}{rgb}{0.21,0.49,0.74}
\usepackage[pagebackref,breaklinks,colorlinks,allcolors=cvprblue]{hyperref}

\def\paperID{91} 
\def\confName{CVPR}
\def\confYear{2026}

\title{GeoSAE: Geometric Prior-Guided Layer-Wise Sparse Autoencoder Annotation of Brain MRI Foundation Models}

\author{Favour Nerrise \quad
Lucy Yin \quad
Mohammad H. Abbasi \quad
Kilian M. Pohl \quad
Ehsan Adeli\thanks{Corresponding author.}\\
Stanford University, CA\\
{\tt\small \{fnerrise, lucyyin, mabbasi, kpohl, eadeli\}@stanford.edu}}

\begin{document}
\maketitle
\begin{abstract}
    Brain MRI foundation models learn rich representations of anatomy, but interpreting what clinical information they encode remains an open problem. Standard sparse autoencoders (SAEs) suffer from severe feature collapse in deep transformer layers, and in Alzheimer's disease (AD) research, aging confounds nearly every clinical variable, making na\"ive annotation unreliable. We propose \paper, a geometry-guided SAE framework that uses the foundation model's learned manifold structure to prevent feature collapse and annotates each surviving feature via age-deconfounded partial correlations. Applied to ${\sim}$14k T1-weighted MRI scans from the Alzheimer's Disease Neuroimaging Initiative (ADNI) and the Australian Imaging Biomarkers and Lifestyle (AIBL) datasets, \paper identifies a compact, fully interpretable feature set that predicts mild cognitive impairment (MCI)-to-AD conversion (AUC 0.746) using only 2\% of the embedding dimensions, while comorbidity-annotated features achieve only chance-level performance. The identified features replicate across cohorts without retraining ($r{=}0.97$) and localize to neuroanatomically distinct regions consistent with Braak staging. This shows that geometry-guided SAEs can extract interpretable, biomarkers from frozen brain MRI foundation models.
    \end{abstract}
    
\section{Introduction}
\label{sec:intro}

\begin{figure*}[t]
    \centering
    \includegraphics[width=\textwidth]{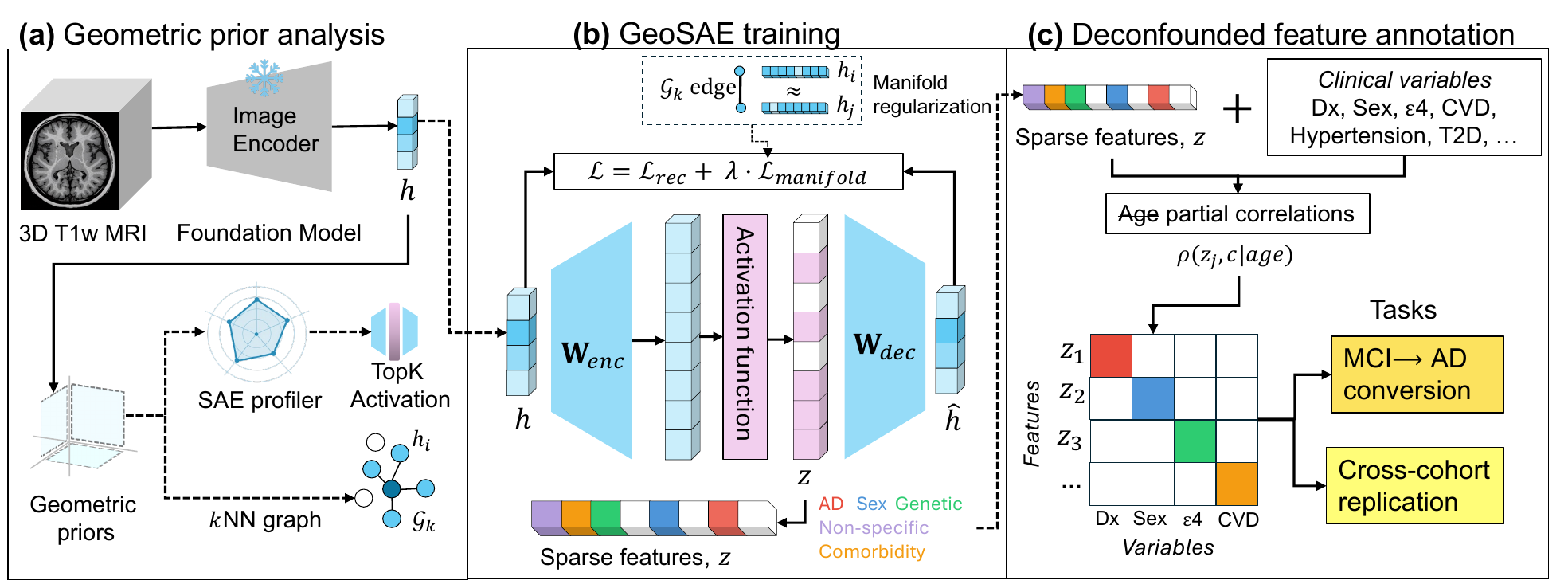}
    \caption{\textbf{Overview of GeoSAE.}
        (a)~Geometric prior analysis of a frozen brain MRI FM selects the SAE activation function and constructs a $k$-NN manifold graph. (b)~GeoSAE training uses manifold regularization to prevent feature collapse. (c)~Age-deconfounded feature annotation assigns each alive feature to a clinical category for downstream tasks.}
    \label{fig:overview}
\end{figure*}

Brain MRI foundation models (FMs) pretrained on tens of thousands of scans via self-supervised learning achieve strong downstream transfer for disease classification, age prediction, and brain-age estimation~\cite{tak2026generalizable,barbano2025anatomical,kaczmarek2025building}. However, these models produce high-dimensional, entangled representations whose clinical content is difficult to interpret. In Alzheimer's disease (AD) research, where aging is both the primary risk factor and a pervasive confound, this opacity poses a direct clinical risk. Features that appear to predict AD progression may instead reflect normal aging, comorbidity burden, or scanner artifacts~\cite{polsterl2023identification}. Reliable deployment of brain MRI FMs therefore requires methods that can decompose their representations into discrete, interpretable features and that can control for age when assigning clinical meaning~\cite{fay2023avoiding,coluzzi2025biomarker}.

Existing interpretation methods neither decompose FM representations into discrete features nor control for confounds. Post-hoc attribution methods such as GradCAM~\cite{selvaraju2017grad} and integrated gradients~\cite{sundararajan2017axiomatic} produce spatial saliency maps that highlight which input voxels influence a prediction, but they do not identify what high-level concepts the model has learned to represent internally. Linear probing~\cite{alain2016understanding} and concept activation vectors~\cite{kim2018interpretability} test whether specific clinical variables are linearly decodable from intermediate representations, but they treat the latent space as a monolithic vector and cannot determine which dimensions encode which information. Disentanglement-based methods~\cite{jung2025disclose,Higgins2016betaVAELB} and concept-based approaches~\cite{gao2025learning} can isolate individual factors of variation, but they require predefined labels during training and architectural modifications to the model itself. Brain MRI FMs are pretrained at scale via self-supervision, so a post-hoc method that can decompose their learned representations into discrete, identifiable features without modifying the model is essential.

Sparse autoencoders (SAEs) provide such a method. An SAE trains a sparse, overcomplete dictionary on a model's internal activations so that each representation is reconstructed from a small set of active dictionary elements, where each element ideally corresponds to a single interpretable concept~\cite{cunningham2023sparse,bricken2023toward}. Researchers have used SAEs to reveal monosemantic features in large language models~\cite{templeton2024scaling,gao2024scaling}, to identify interpretable directions in vision transformers~\cite{olson2025probing,pach2025sparse}, and to discover biologically meaningful structure in protein language models~\cite{simon2025interplm,garcia2025interpreting}. Recent extensions to medical imaging FMs have shown promise in pathology~\cite{le2024learning}, radiology~\cite{abdulaal2024x}, and hematology~\cite{dasdelen2025cytosae}. However, all of these applications operate on models where feature mortality is modest and domain-specific confounds do not arise. Applying SAEs to neuroimaging FMs introduces three distinct challenges: \textit{(a)}~standard SAEs suffer severe feature mortality in deep transformer layers, losing $>$98\% of dictionary elements because no geometric prior guides regularization; \textit{(b)}~age confounds nearly every clinical variable, so raw feature-outcome correlations conflate neurodegeneration with normal aging; and \textit{(c)}~comorbidities such as cardiovascular disease and diabetes share risk factors with AD, so feature annotations must disentangle disease-specific signal from comorbidities.

\noindent\textbf{Our approach.} We address all three challenges with \paper (\textbf{Geo}metry-guided \textbf{S}parse \textbf{A}uto\textbf{E}ncoder), a framework that we apply to a 3D brain MRI ViT~\cite{tak2026generalizable} (Fig.~\ref{fig:overview}). \paper first analyzes the geometry of each transformer layer's representations to select an appropriate SAE activation function and to construct a $k$-NN manifold graph. To address feature mortality~\textit{(a)}, we introduce a manifold regularization term that penalizes dissimilar pre-activations for samples that are neighbors on the FM's learned manifold. This regularization supplies gradients to all dictionary elements, including those zeroed out by the sparsity gate, and thereby prevents feature collapse (Sec.~\ref{sec:geosae}). To address age confounding~\textit{(b)} and comorbidity confounding~\textit{(c)}, we annotate each surviving feature via age-deconfounded partial correlations with hierarchical FDR correction and validate the annotations against 40+ secondary clinical variables.

We evaluate \paper on approximately 14,500 T1-weighted MRI scans from ADNI~\cite{jack2008alzheimer} and AIBL~\cite{ellis2009australian}. At the optimal layer (layer~9), \paper produces 161 alive features, $7{\times}$ more than a standard SAE. The top 16 of these features achieve 100\% clinical annotation and predict MCI-to-AD conversion with AUC 0.749, outperforming the full 768-dimensional CLS representation (0.732) while providing clinically interpretable features that replicate across cohorts ($r{=}0.97$). Cross-cohort activation consistency reaches $r{=}0.97$, a 32\% improvement over standard SAEs ($r{=}0.74$). Attention-based saliency maps localize each feature to brain regions consistent with Braak staging~\cite{therriault2022biomarker}.

Our contributions are threefold: (1)~We propose \paper, which uses geometric analysis of FM representations to guide SAE activation function selection and introduces a manifold regularization on pre-activations that prevents feature collapse. (2)~We design an age-deconfounded annotation pipeline using partial correlations with hierarchical FDR correction to assign each surviving feature to a clinical category. (3)~We apply \paper across all 12 transformer layers of a brain MRI FM and show that 16 interpretable features at the peak layer predict AD conversion better than the full representation, replicate across independent cohorts, and localize to neuroanatomically plausible regions.\footnote{Code: \url{https://github.com/favour-nerrise/GeoSAE}}

\section{Related Works}
\label{sec:related}

\noindent\textbf{Interpreting neural network representations.}
The problem of understanding what a neural network has learned has motivated a broad family of methods. Gradient-weighted class activation mapping (GradCAM)~\cite{selvaraju2017grad} computes class-discriminative saliency by weighting feature maps with gradients, while integrated gradients~\cite{sundararajan2017axiomatic} accumulate gradients along a path from a baseline input. SHAP~\cite{lundberg2017unified} unifies several attribution approaches under a game-theoretic framework. All of these methods produce per-voxel importance scores and are effective for localizing predictions, but they do not decompose the model's internal representation into discrete, nameable concepts. Linear probing~\cite{alain2016understanding} addresses this by training linear classifiers on intermediate representations to test whether specific variables are encoded, and concept activation vectors (TCAVs)~\cite{kim2018interpretability} quantify a model's sensitivity to user-defined concept directions. Both approaches, however, treat the latent space as a single vector and cannot identify which dimensions or subspaces encode which information. Disentanglement methods such as $\beta$-VAE~\cite{Higgins2016betaVAELB} and supervised disentanglement~\cite{jung2025disclose,wang2024disentangled} can separate individual factors of variation, but they require labels during training and architectural modifications. Concept bottleneck models~\cite{gao2025learning} similarly require concept supervision. None of these methods can be applied post-hoc to a frozen, self-supervised FM. Our work uses SAEs, which require no labels and no model modification, to decompose frozen FM representations into interpretable features.

\noindent\textbf{Sparse autoencoders for mechanistic interpretability.}
SAEs were first applied to language models, where Bricken et al.~\cite{bricken2023toward} demonstrated that a one-layer MLP in a small transformer could be decomposed into monosemantic dictionary elements. Templeton et al.~\cite{templeton2024scaling} scaled this approach to Claude~3 Sonnet and showed that SAE features correspond to human-interpretable concepts at the level of individual neurons. Gao et al.~\cite{gao2024scaling} systematically evaluated SAE scaling laws and training recipes. In vision, Olson et al.~\cite{olson2025probing} applied SAEs to CLIP and DINOv2 and found that features align with object parts and textures, Pach et al.~\cite{pach2025sparse} showed that SAE features in vision-language models are monosemantic, and Stevens et al.~\cite{stevens2025interpretable} proposed evaluation criteria for scientifically rigorous SAE interpretation. Lim et al.~\cite{lim2024sparse} introduced PatchSAE, which extracts interpretable concepts with patch-wise spatial attributions from CLIP and reveals how adaptation techniques remap visual concepts. In the biomedical domain, Le et al.~\cite{le2024learning} used SAEs on a pathology FM and annotated features via enrichment testing against gene expression programs. Abdulaal et al.~\cite{abdulaal2024x} applied SAEs to a radiology FM for report generation, and Dasdelen et al.~\cite{dasdelen2025cytosae} decomposed a hematology FM into cell-type-specific features. Simon and Zou~\cite{simon2025interplm} applied SAEs to protein language models and annotated features against Gene Ontology terms. A common finding across these studies is that SAE features are more interpretable and more specific than individual neurons. However, none of these works address feature mortality in deep layers or control for confounds during annotation. We address both with geometric regularization and age-deconfounded partial correlations.

\noindent\textbf{Brain MRI foundation models.}
Self-supervised pretraining on large-scale structural MRI has produced FMs with strong transfer to clinical tasks. BrainIAC~\cite{tak2026generalizable} trains a ViT on over 85,000 brain MRI scans and achieves state-of-the-art performance on age prediction and disease classification. Barbano et al.~\cite{barbano2025anatomical} pretrain anatomical FMs that capture regional brain structure, Kaczmarek et al.~\cite{kaczmarek2025building} build a SimCLR-based FM for neurological disease diagnosis from 3D MRI, and Rui et al.~\cite{rui2025multi} propose multi-view pre-training that jointly learns from multiple brain imaging modalities. Wald et al.~\cite{wald2025revisiting} revisit MAE pre-training for 3D medical image segmentation, demonstrating the effectiveness of self-supervised ViTs for volumetric data. These models learn rich representations of brain anatomy, but prior work has shown that neuroimaging embeddings frequently encode scanner manufacturer and acquisition site alongside clinical signal~\cite{loaliyan2024comparative}, that aging and neurodegeneration produce overlapping structural changes~\cite{polsterl2023identification}, and that comorbidities such as cardiovascular disease and diabetes further confound disease-specific signatures~\cite{fay2023avoiding,coluzzi2025biomarker}. No prior work has attempted to decompose a brain MRI FM's representations into interpretable features while controlling for these confounds. Our work is the first to apply SAEs to brain MRI FMs and to use the FM's own geometric structure to regularize SAE training.

\section{Methods}
\label{sec:methods}

\paper is a three-stage pipeline (Fig.~\ref{fig:overview}): geometric prior analysis selects an activation function and constructs a manifold graph (Sec.~\ref{sec:background}); a manifold-regularized SAE uses this graph to prevent feature collapse (Sec.~\ref{sec:geosae}); and age-deconfounded annotation assigns each surviving feature to a clinical category (Sec.~\ref{sec:annotation}).

\subsection{Sparse Autoencoders for Brain MRI Foundation Models}
\label{sec:background}

\noindent\textbf{Background.}
Let $\cD = \{(\bx_i, \mathbf{c}_i)\}_{i=1}^N$ denote $N$ T1-weighted brain MRI scans with clinical covariates $\mathbf{c}_i = (\text{age}, \text{sex}, \text{APOE4}, \text{diagnosis}, \mathbf{m})$, where $\mathbf{m} \in \{0,1\}^M$ are $M$ comorbidity indicators. A frozen, self-supervised FM produces per-layer [CLS] token representations $\bh_i^{(\ell)} = f_\text{FM}^{(\ell)}(\bx_i) \in \R^{d}$ for layers $\ell = 1, \ldots, L$.
A sparse autoencoder (SAE)~\cite{cunningham2023sparse} maps each dense $\bh \in \R^d$ to a sparse code $\bz \in \R^{d_\text{SAE}}$ ($d_\text{SAE} = d \times E$, expansion factor $E$). The encoder computes pre-activations $\ba = \bW_\text{enc}(\bh - \bb_\text{pre}) + \bb_\text{enc}$ and applies a sparsity-inducing activation $\bz = \sigma(\ba)$. The decoder reconstructs $\hat{\bh} = \bW_\text{dec}\,\bz + \bb_\text{pre}$ with unit-norm columns, minimizing $\mathcal{L}_\text{SAE} = \|\bh - \hat{\bh}\|_2^2$. Each active dimension of $\bz$ ideally corresponds to one interpretable concept.

\noindent\textbf{Feature mortality in deep layers.}
SAEs have been successfully applied to language models~\cite{templeton2024scaling,gao2024scaling} and vision transformers~\cite{olson2025probing}, but brain MRI FMs pose a distinct challenge. A feature $j$ is \emph{alive} if $\exists\, i : z_{ij} > 0$ and \emph{dead} otherwise. As transformer representations become increasingly structured with depth, most dictionary elements receive zero gradient through the sparsity gate and are never updated. Dead neuron resampling~\cite{bricken2023toward} partially addresses this but destabilizes training for larger dictionaries. We propose instead to use the FM's own geometric structure to regularize SAE training.

\noindent\textbf{Geometric prior analysis.}
We select the SAE activation function by evaluating four candidates (ReLU~\cite{bricken2023toward}, JumpReLU~\cite{rajamanoharan2024jumping}, TopK~\cite{makhzani2013k}, SpaDE~\cite{hindupur2025projecting}) against five geometric properties of each layer's representations, including angular vs.\ radial class structure, dimensionality homogeneity, and sparsity uniformity. For BrainIAC, all layers exhibit angular-dominant structure ($\eta^2{=}0.002$) with significant negative activations; four of five properties favor TopK.

\noindent\textbf{Manifold graph construction.}
Beyond activation selection, the geometric analysis yields a $k$-nearest-neighbor manifold graph $\mathcal{N}_k$ on the FM embeddings. We compute pairwise distances among all $N$ representations $\{\bh_i\}$ and connect samples to $k$ nearest neighbors with Gaussian kernel weights:
\begin{equation}
    w_{ij} = \exp\!\Big(-\frac{\|\bh_i - \bh_j\|^2}{2\sigma^2}\Big)
    \label{eq:knn}
\end{equation}
where $\sigma$ is the median neighbor distance. Samples that the FM considers similar (nearby in representation space) receive high edge weights. The graph serves as the manifold prior for \paper training and provides the neighborhood structure that guides regularization in Sec.~\ref{sec:geosae}.

\subsection{Manifold-Regularized Sparse Autoencoder}
\label{sec:geosae}

\noindent\textbf{TopK activation.}
In our dataset, geometric prior analysis recommends TopK~\cite{makhzani2013k} as the activation function across all 12 layers. We train $f_\text{SAE}^{(\ell)}$ independently for each layer~$\ell$. TopK activation retains only the $k$ largest positive pre-activation entries:
\begin{equation}
    z_j = \begin{cases}
        \text{ReLU}(a_j) & \text{if } j \in \text{top-}k(\ba) \\
        0 & \text{otherwise}
    \end{cases}
    \label{eq:topk}
\end{equation}
TopK enforces exact sparsity by activating precisely $k$ features per input. This simplifies comparison across layers and avoids sensitivity to the $\ell_1$ penalty weight. Features outside the top-$k$ set receive exactly zero gradient and remain dead.

\noindent\textbf{Manifold regularization.}
\paper addresses feature mortality by adding a manifold smoothness penalty on \emph{pre}-activations $\ba$ (before the sparsity gate) using the $k$-NN graph $\mathcal{N}_k$ from Sec.~\ref{sec:background}:
\begin{equation}
    \mathcal{L} = \mathcal{L}_\text{SAE} + \lambda
    \sum_{(i,j) \in \mathcal{N}_k} w_{ij}\,
    \|\ba_i - \ba_j\|_2^2
    \label{eq:manifold}
\end{equation}
This penalty operates on pre-activations, not post-activation codes. Every encoder-based SAE computes pre-activations before applying a sparsity gate, so the manifold term provides gradients to \emph{all} dictionary elements, including those zeroed out by TopK. Samples that are neighbors on the FM's learned manifold are encouraged to have similar pre-activation patterns. This distributes gradient signal across the full dictionary and prevents collapse to a small active set.

\noindent\textbf{Properties.}
The manifold term operates on pre-activations and is therefore activation-agnostic; it can be combined with any sparsity gate. Setting $\lambda{=}0$ recovers the standard SAE and provides a natural ablation baseline. Subsequent analyses are restricted to alive features ($\exists\, i : z_{ij} > 0$).

\noindent\textbf{Comparison with dead neuron resampling.}
The standard remedy for feature mortality is dead neuron resampling~\cite{bricken2023toward}, which periodically reinitializes dead dictionary elements. In our setting, resampling has two drawbacks. First, it causes training instability when combined with manifold regularization because resampled features are initialized far from the manifold-consistent landscape. Second, resampling alone produces features that converge to the dominant signal rather than diverse clinical categories. Manifold regularization avoids both issues because it provides continuous gradient flow to all dictionary elements. The $k$-NN graph is precomputed once ($\mathcal{O}(N^2 d)$); the per-batch manifold loss adds $\mathcal{O}(Bk\,d_\text{SAE})$ operations.

\subsection{Deconfounded Feature Annotation}
\label{sec:annotation}

\noindent\textbf{Age confounding.}
We annotate each alive feature by its association with $3 + M$ clinical variables: age, sex, APOE4, diagnosis, and $M$ comorbidities. Prior SAE annotation methods~\cite{le2024learning,simon2025interplm,dasdelen2025cytosae} do not adjust for covariates. In neuroimaging, this confounds aging with neurodegeneration. Without correction, nearly every feature appears ``AD-related'' because both the feature and diagnosis correlate with age.

\noindent\textbf{Partial correlations.}
We compute partial Spearman correlations controlling for age. For variable $c$ and feature $j$, let $r_{jc}$, $r_{j,a}$, and $r_{c,a}$ denote Spearman correlations between the respective rank-transformed quantities and age~$a$:
\begin{equation}
    \rho_{jc \cdot a} = \frac{r_{jc} - r_{j,a}\,r_{c,a}}
    {\sqrt{1 - r_{j,a}^2}\;\sqrt{1 - r_{c,a}^2}}
    \label{eq:partial}
\end{equation}
This removes the variance shared with age from both the feature and the clinical variable before computing their association. Testing all feature$\times$variable pairs simultaneously inflates false positives, so we apply false discovery rate (FDR) correction~\cite{benjamini1995controlling} to $p$-values from the partial-correlation matrix.

\noindent\textbf{Category assignment.}
Let $\mathcal{C}$ denote the $3 + M$ non-age variables, grouped into categories (AD-related, sex-related, genetic, comorbidity). Each alive feature~$j$ is assigned to the category of the variable with the strongest significant partial correlation:
\begin{equation}
    \text{cat}(j) = \arg\max_{c \in \mathcal{C}}\;
    |\rho_{jc \cdot a}| \quad \text{s.t.}\;\;
    p_{jc}^{\,\text{FDR}} < 0.05
    \label{eq:category}
\end{equation}
We label features with no significant partial correlation but significant raw age correlation ($p_\text{age}^{\text{FDR}} < 0.05$) as ``aging'' and the remainder as ``non-specific.'' This ensures that each feature receives a single, interpretable clinical label while respecting the dominant confound in neuroimaging data.

\section{Experiments}

\subsection{Dataset and Setup}
\label{sec:setup}

\noindent\textbf{Datasets.} For training and primary evaluation, we use T1-weighted MRI from the Alzheimer's Disease Neuroimaging Initiative (ADNI) dataset~\cite{jack2008alzheimer}, consisting of 13{,}218 scans across 1{,}974 participants (34.6\% cognitively normal (CN), 53.1\% MCI, 12.4\% AD; mean age 74.6$\pm$7.4; 46.9\% female). We also use the Australian Imaging Biomarkers and Lifestyle Study of Ageing (AIBL) dataset~\cite{ellis2009australian}, which includes 1{,}266 scans across 690 participants (67.9\% CN (including 7.6\% subjective memory complaints), 13.5\% MCI, 11.0\% AD; mean age 73.7$\pm$6.7; 53.1\% female), for external validation. We preprocessed all volumes~\cite{abbasi2025smri} with skull-stripping, bias-correction, registration to MNI space, and trilinear resampling to $96^3$. Within ADNI, we derive MCI-to-AD conversion labels from longitudinal diagnostic records (926 MCI subjects: 343 converters, 583 stable); AIBL is used for feature annotation replication only, as its MCI converter count ($n{=}11$) is insufficient for conversion evaluation. Clinical covariates include $M{=}5$ comorbidity indicators in ADNI (hypertension, hyperlipidemia, depression, type~2 diabetes, cardiovascular disease) and $M{=}4$ in AIBL (hypertension unavailable). ADNI additionally provides 31 secondary variables spanning demographics, vitals, psychiatric history, cardiovascular indicators, metabolic markers, lifestyle factors, medications, and scanner metadata.

\noindent\textbf{Evaluation tasks.} We validate the annotated features through three tasks. \textit{(1)~Enrichment testing} (ADNI): we test each category against the 31 secondary variables using per-variable FDR correction. AD-related features should be enriched for AD-adjacent variables (e.g.\ cognitive scores, AD medications); we characterize features with no clinical association by activation frequency and image-quality statistics to assess whether they encode data-quality artifacts. \textit{(2)~Selective prediction} (ADNI): we evaluate category-restricted feature subsets for MCI-to-AD conversion prediction via logistic regression (5-fold stratified CV). The key test is that AD-related features should predict conversion comparably to all features, while comorbidity features should not. \textit{(3)~Cross-cohort replication} (AIBL): we apply the ADNI-trained SAE to AIBL without retraining and recompute all clinical correlations. We quantify replicability by annotation agreement (Pearson $r$ between feature-level correlations across cohorts) and activation consistency (Spearman $r$ between per-feature activation magnitudes).

\noindent\textbf{Implementation.} The FM is BrainIAC~\cite{tak2026generalizable}, a 12-layer ViT ($L{=}12$, $d{=}768$). For cross-layer analysis, we train $L$ \paper models ($\lambda{=}0.1$, $k_\text{NN}{=}15$, one per layer) with $E{=}2$, $k{=}16$ ($d_\text{SAE}{=}1{,}536$). At the peak layer (layer~9), we also train a standard SAE ($\lambda{=}0$) for comparison: it produces 23 alive features versus 161 for \paper ($7{\times}$). We train all SAEs for 100 epochs with Adam (lr $10^{-3}$, batch 256) on a 90/10 subject-level split, achieving $>$99\% explained variance. For external validation, we apply the same layer-9 \paper to AIBL without retraining. For single-scan prediction, we use the latest available scan per subject to ensure one observation per subject. All conversion experiments use 5-fold stratified cross-validation with subject-level splits (all scans from a subject appear in the same fold) to prevent data leakage from longitudinal scans.

\begin{figure}[t]
    \centering
    \includegraphics[width=\linewidth]{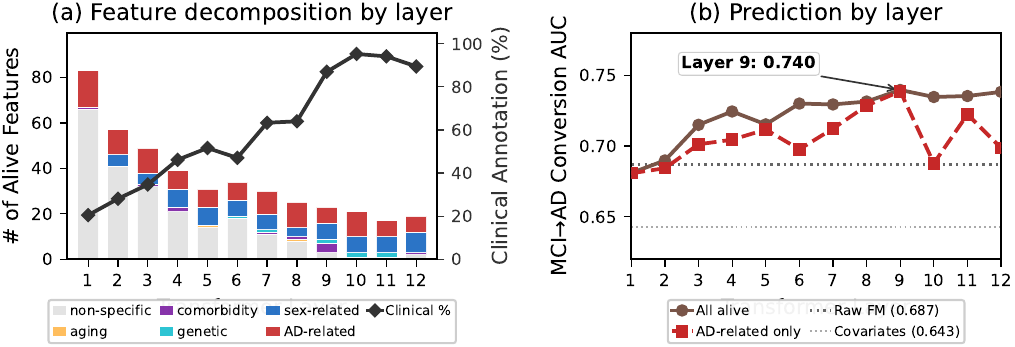}
    \caption{Cross-layer analysis of \paper across 12 BrainIAC layers. (a)~Stacked bars show alive features by clinical category; the line shows clinical annotation rate. Features consolidate with depth while clinical specificity increases. (b)~Conversion AUC peaks at layer~9, then declines as scanner features dominate.}
    \label{fig:crosslayer}
    \end{figure}

\subsection{Cross-Layer Decomposition}
\label{sec:results_derive}

\noindent\textbf{Geometric prior results.}
Geometric prior analysis of the BrainIAC representations identifies angular-dominant structure ($\eta^2{=}0.002$), homogeneous dimensionality, uniform sparsity, and significant negative activations (48.8\%); four of five geometric properties favor TopK, and the analysis rules out simplex-based methods. These properties are consistent across all 12 layers, so we use TopK as the activation function for all \paper models.

\noindent\textbf{Layer-wise trends.}
Training \paper independently at each of the 12 transformer layers reveals two trends (Fig.~\ref{fig:crosslayer}). First, alive features peak in early layers (447 at layer~2) and consolidate with depth (147 at layer~10), a pattern that mirrors the FM's progressive abstraction from low-level texture to high-level semantic content. Second, conversion AUC increases with depth and peaks at layer~9 (0.731$\pm$0.011), then declines as scanner-related features dominate layers~10--12.

\begin{figure}[t]
\centering
\includegraphics[width=0.85\linewidth]{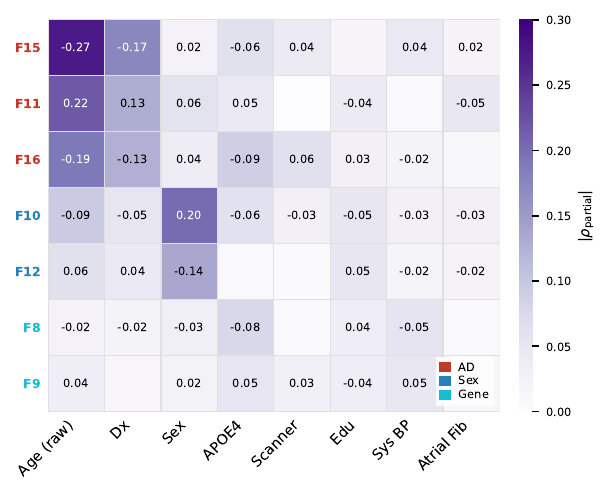}
\caption{\paper annotation (layer~9).
$|\rho_{jc \cdot a}|$ for 3 strongest per category (\textcolor{adred}{AD}, \textcolor{sexblue}{Sex}, \textcolor{geneteal}{Genetic});
values where FDR $p{<}0.05$.}
\label{fig:heatmap}
\end{figure}

\noindent\textbf{Clinical annotation across layers.}
Clinical annotation rates range from 14\% to 29\% across layers. At layer~9, the top-16 features by activation frequency achieve 100\% annotation (Fig.~\ref{fig:heatmap}): 5 AD-related, 6 sex-related, and 5 genetic features, with zero non-specific. Non-specific features dominate early layers (80\% at layer~1), while AD, sex, and genetic features account for 87\% at layer~9. This progressive specialization indicates that deeper layers encode increasingly clinical, rather than low-level, information.

\subsection{Selective Prediction}
\label{sec:results_prediction}

\noindent\textbf{Feature characterization.}
At layer~9, \paper training ($\lambda{=}0.1$) produces 161 alive features ($7{\times}$ more than standard) with $17{\times}$ lower inter-feature redundancy (mean pairwise $|r|{=}0.009$ vs.\ $0.156$). The top-16 features by activation frequency achieve 100\% clinical annotation (5~AD, 6~sex, 5~genetic; zero non-specific) and AUC 0.746.

\noindent\textbf{Single-scan prediction.}
Table~\ref{tab:main} presents single-scan MCI-to-AD conversion results. \paper (16 features) achieves the highest AUC (0.746), surpassing both a supervised 3D ResNet-18~\cite{he2016deep} baseline (MONAI~\cite{cardoso2022monai}) trained from scratch on the same MRI volumes (0.729) and the raw 768-dim FM embedding (0.687), while using only 16 dimensions ($48{\times}$ fewer than the raw embedding), the same reduction as PCA and $\beta$-VAE but with fully interpretable, cross-cohort-stable features ($r{=}0.97$ vs.\ $0.77$; Sec.~\ref{sec:results_cross}). We also include \paper-rand, a control that samples 16 alive \paper features uniformly at random rather than selecting the top-16 by activation frequency; its near-chance performance (0.574) shows that the gain does not come from arbitrary feature subsampling. PCA and $\beta$-VAE~\cite{Higgins2016betaVAELB} at the same dimensionality ($d{=}16$) achieve 0.734 and 0.740 respectively, but the $\beta$-VAE shows substantially lower cross-cohort replicability ($r{=}0.77$ vs.\ $0.98$; Sec.~\ref{sec:results_cross}). The standard SAE (23 alive features, no manifold regularization) achieves 0.737; \paper surpasses it with fewer features, which demonstrates the benefit of manifold-guided feature selection. All imaging methods outperform non-imaging covariates (0.643), which confirms that the FM encodes conversion-relevant structure beyond age, sex, and APOE4.

\begin{table}[t]
\centering
\caption{Single-scan MCI-to-AD conversion prediction (5-fold stratified CV, 926 MCI subjects: 343 converters, 583 stable). $d$: input dimensionality. Sens/Spec at threshold 0.5. \textbf{Bold}: best; \underline{underline}: second best.}
\label{tab:main}
\footnotesize
\begin{tabular}{@{}l@{\hskip 4pt}c@{\hskip 4pt}c@{\hskip 4pt}c@{\hskip 4pt}r@{}}
\toprule
\textbf{Model} & \textbf{$d$} & \textbf{AUC} & \textbf{Sens (\%)} & \textbf{Spec (\%)} \\
\midrule
Non-imaging (covariates) & 3 & .643\,$\pm$\,.003 & 24.1\,$\pm$\,0.4 & \underline{85.8}\,$\pm$\,0.5 \\
3D ResNet-18~\cite{he2016deep} & $96^3$ & .729\,$\pm$\,.035 & \textbf{66.0}\,$\pm$\,20.0 & 61.4\,$\pm$\,18.4 \\
Raw FM~\cite{tak2026generalizable} & 768 & .687\,$\pm$\,.009 & \underline{54.0}\,$\pm$\,2.3 & 71.6\,$\pm$\,0.7 \\
PCA~\cite{joliffe1992principal} & 16 & .734\,$\pm$\,.004 & 47.0\,$\pm$\,1.1 & 82.9\,$\pm$\,0.6 \\
$\beta$-VAE~\cite{Higgins2016betaVAELB} & 16 & \underline{.740}\,$\pm$\,.002 & 48.8\,$\pm$\,0.9 & 83.7\,$\pm$\,0.6 \\
\midrule
ReLU SAE~\cite{bricken2023toward} & 16 & .728\,$\pm$\,.003 & 45.9\,$\pm$\,1.3 & 81.9\,$\pm$\,0.7 \\
TopK SAE~\cite{makhzani2013k} & 16 & .737\,$\pm$\,.003 & 47.9\,$\pm$\,0.8 & 83.9\,$\pm$\,0.6 \\
\midrule
\paper-rand & 16 & .574\,$\pm$\,.064 & 11.0\,$\pm$\,12.8 & \textbf{94.1}\,$\pm$\,6.1 \\
\paper (ours) & 16 & \textbf{.746}\,$\pm$\,.002 & \underline{49.2}\,$\pm$\,0.6 & 82.1\,$\pm$\,0.7 \\
\bottomrule
\end{tabular}
\end{table}

\subsection{Ablation Studies}
\label{sec:ablation}

\noindent\textbf{Component ablations.}
Table~\ref{tab:ablation} ablates GeoSAE's key components. Removing manifold regularization ($\lambda{=}0$) reduces alive features from 161 to 23 and annotation from 100\% to 87\%, while over-regularization ($\lambda{=}10$) degrades AUC. AUC is stable across $\lambda \in [0, 5]$ ($\Delta$AUC $< 0.002$ except $\lambda{=}0.5$) and similarly robust to the manifold graph neighborhood size ($k_\text{NN} \in \{5, 10, 15, 20, 30\}$, $\Delta$AUC $< 0.006$). Sweeping dictionary size and sparsity ($E \in \{2,4,8\}$, $k \in \{8,16,32\}$) confirms that $E{=}2$, $k{=}16$ is optimal; larger dictionaries fragment features and reduce AUC. Without age deconfounding, zero features are assigned to AD-related because $|\rho_\text{age}|{>}|\rho_\text{diagnosis}|$ for every feature.

\noindent\textbf{Category ablation.}
Category-level feature removal tests whether annotations reflect genuine predictive contributions. Dropping AD-related features causes the largest AUC decrease ($-$0.017), while dropping comorbidity features has negligible effect ($-$0.003), confirming that AD features are both sufficient (0.720 alone) and \emph{necessary} for conversion prediction. Dropping the 122 non-specific features \emph{improves} AUC to 0.749, which indicates they add noise.

\begin{table}[t]
\centering
\caption{Ablation study. Each row modifies one component of \paper ($E{=}2$, $k{=}16$, $\lambda{=}0.1$). Alive: features with $>$0 activation. \textbf{Bold}/\underline{underline}: best/second-best among ablation variants; category-only rows are negative controls, not competing methods.}
\label{tab:ablation}
\footnotesize
\begin{tabular}{@{}lccccc@{}}
\toprule
\textbf{Variant} & \textbf{Alive} & \textbf{$d$} & \textbf{AUC} & \textbf{Sens (\%)} & \textbf{Spec (\%)} \\
\midrule
\paper (ours) & 161 & 16 & \textbf{.746} & \textbf{49.2} & 82.1 \\
\midrule
\multicolumn{6}{@{}l}{\emph{Manifold regularization ($\lambda$)}} \\[2pt]
\quad w/o manifold ($\lambda{=}0$) & 23 & 16 & .737 & 47.9 & \textbf{83.9} \\
\quad $\lambda{=}1.0$ & 121 & 16 & .740 & 47.3 & \underline{83.7} \\
\quad $\lambda{=}10.0$ & 14 & 16 & .722 & 44.9 & 82.4 \\
\midrule
\multicolumn{6}{@{}l}{\emph{Architecture ($E$, $k$)}} \\[2pt]
\quad $E{=}2$, $k{=}8$ & 156 & 8 & \underline{.743} & 45.7 & 82.3 \\
\quad $E{=}2$, $k{=}32$ & 86 & 32 & .729 & \underline{48.5} & 81.1 \\
\quad $E{=}4$, $k{=}16$ & 383 & 16 & .734 & 46.3 & 82.9 \\
\quad $E{=}8$, $k{=}16$ & 1404 & 16 & .735 & 45.9 & 82.8 \\
\midrule
\multicolumn{6}{@{}l}{\emph{Category-only features}} \\[2pt]
\quad AD-related only & 7 & 7 & .720 & 43.8 & 81.2 \\
\quad Comorbidity only & 13 & 13 & .507 & 0.4 & 99.7 \\
\bottomrule
\end{tabular}
\end{table}

\begin{figure*}[t]
\centering
\includegraphics[width=\textwidth]{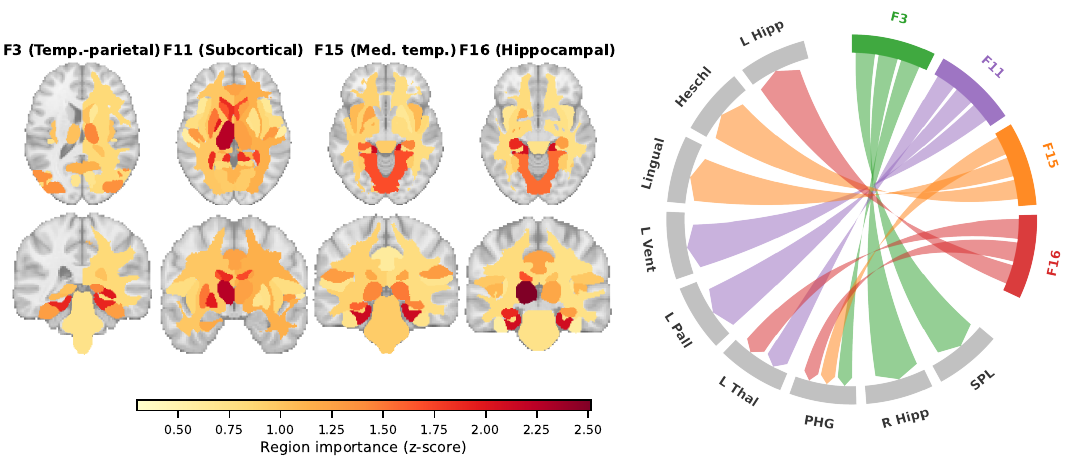}
\caption{Brain region localization of the top-4 conversion-predictive \paper features (F3, F11, F15, F16: top-4 SAE features ranked by individual MCI-to-AD conversion AUC) via attention rollout (layers 1--9) mapped to the Harvard-Oxford atlas. Left: axial (top) and coronal (bottom) slices; color intensity indicates z-scored region importance. Right: chord diagram showing top-3 regions per feature. PHG: parahippocampal gyrus; SPL: superior parietal lobule; L/R Hipp: hippocampus; L Thal/Pall/Vent: thalamus, pallidum, lateral ventricle.}
\label{fig:brainmaps}
\end{figure*}

\subsection{Brain Region Localization}
\label{sec:results_brain}

\noindent\textbf{Attention rollout.}
Clinical annotation identifies \emph{what} each SAE feature encodes; we next ask \emph{where} in the brain it attends. We compute attention rollout through the first 9 ViT layers, aggregate over the 50 maximally activating samples per feature, and map results to the Harvard-Oxford atlas~\cite{jenkinson2012fsl}.

\noindent\textbf{Anatomical sub-patterns.}
The top-4 features ranked by individual MCI-to-AD conversion AUC localize to anatomically distinct sub-patterns (Fig.~\ref{fig:brainmaps}). Features F1 and F2 attend primarily to medial temporal lobe structures (hippocampus, parahippocampal gyrus), consistent with early Braak stages where tau pathology first appears~\cite{therriault2022biomarker}. Feature F3 targets subcortical nuclei (thalamus, pallidum, ventricles), which reflect global atrophy and ventricular enlargement characteristic of advancing neurodegeneration~\cite{planche2022structural}. Feature F4 activates a temporo-parietal network (superior parietal lobule, amygdala, fusiform cortex), consistent with later Braak stages where pathology spreads to association cortices~\cite{therriault2022biomarker}.

\noindent\textbf{Visualization.}
Fig.~\ref{fig:brainmaps} shows axial and coronal slices with z-scored region importance overlaid on a template brain. The spatial patterns are visually distinct across features: F1--F2 highlight bilateral medial temporal regions, F3 shows a midline subcortical pattern, and F4 produces a distributed lateral pattern. The chord diagram (right panel) confirms that the top-3 regions per feature do not overlap, which indicates that \paper decomposes the FM's conversion prediction into neuroanatomically separable components.

\begin{figure}[t]
    \centering
    \includegraphics[width=0.75\linewidth]{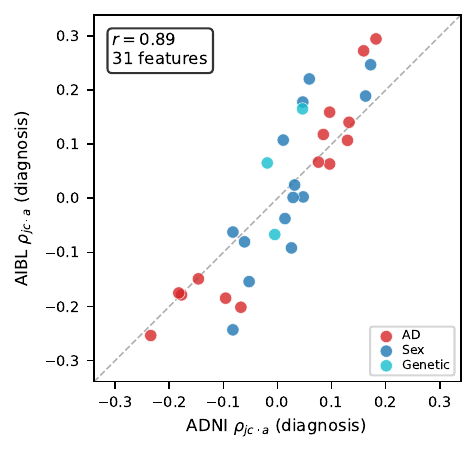}
    \caption{Cross-cohort replication of \paper feature annotations. Each point is an SAE feature alive in both ADNI and AIBL, colored by clinical category. Age-partial diagnosis correlations replicate strongly ($r{=}0.89$, $p{<}10^{-10}$) despite different cohort variables.}
    \label{fig:scatter}
\end{figure}

\subsection{Cross-Cohort Replication and Enrichment}
\label{sec:results_cross}

\noindent\textbf{Annotation agreement.}
When we apply the ADNI-trained layer-9 \paper to AIBL without retraining, it shows strong annotation agreement (Fig.~\ref{fig:scatter}): $r{=}0.89$ for age-partial diagnosis correlations and $r{=}0.84$ for age, with consistent category distributions (7 vs.\ 6 AD-related) despite AIBL's different demographics (75\% CN vs.\ 35\% in ADNI, higher cardiovascular disease and type~2 diabetes prevalence).

\noindent\textbf{Activation consistency.}
The top-16 features achieve activation magnitude $r{=}0.97$ and diagnosis-specific pattern $r{=}0.98$ (CN) / $0.99$ (AD) between ADNI and AIBL. The standard SAE achieves $r{=}0.74$ / $0.75$ / $0.97$ on the same metrics, a 32\% lower cross-cohort reproducibility. All 16 \paper features replicate in AIBL (100\% replication rate). This strong transfer without retraining demonstrates that the manifold-regularized features capture stable biological signal rather than cohort-specific artifacts.

\noindent\textbf{Enrichment testing.}
Enrichment testing against 31 secondary variables identifies AD medication use as the most prominently encoded secondary signal: memantine (26 of 161 features), donepezil (17), and the composite AD-medication indicator (20). This is consistent with known cholinesterase-inhibitor effects on brain structure~\cite{diaz2023differential}, where medicated patients show different atrophy patterns than unmedicated patients. Scanner manufacturer is associated with 11--29 features per layer, which confirms that acquisition confounds are encoded in the FM's representations. Scanner features concentrate in layers~10--12, which further supports the selection of layer~9 as optimal.

\section{Conclusion}
\label{sec:conclusion}
We present \paper, a geometric prior-guided sparse autoencoder framework for interpreting brain MRI foundation models. \paper addresses feature mortality in deep transformer layers through manifold regularization on pre-activations and produces $7{\times}$ more alive features. A core set of 16 features with 100\% clinical annotation predicts MCI-to-AD conversion (AUC 0.746), while comorbidity-annotated features achieve only chance-level performance (AUC 0.506); this contrast confirms that the framework separates disease from comorbidity signal. These features replicate across cohorts without retraining ($r{=}0.97$) and localize to neuroanatomically distinct sub-patterns aligned with Braak staging. We evaluate on a single FM (BrainIAC) and a single clinical domain (AD), so whether these results extend to other architectures and diseases remains to be established; the $k$-NN graph construction also scales quadratically, though in practice it is a one-time cost ($<$1 minute for 13k scans). More broadly, \paper demonstrates that the geometric structure of foundation model representations can guide the extraction of clinically transparent, mechanistically interpretable features from black-box embeddings.

\section*{Acknowledgements}
This work was supported in part by National Institutes of Health grants AG089169, AG084471, AG073356, DA057567, and AA021697, and the Stanford Institute for Human-Centered AI (HAI) Hoffman-Yee Award. F.N. was supported by the Stanford Graduate Fellowship, Stanford NeuroTech Graduate Training Program Fellowship, Stanford EDGE Fellowship, and Stanford Pathways to Neuroscience Fellowship.
{
    \small
    \bibliographystyle{ieeenat_fullname}
    \bibliography{references}
}


\end{document}